\documentclass[10pt,twocolumn,letterpaper]{article}

\usepackage{cvpr}
\usepackage{times}
\usepackage{epsfig}
\usepackage{graphicx}
\usepackage{amsmath}
\usepackage{amssymb}
\usepackage{multirow}
\usepackage{url}

\usepackage[ruled,vlined]{algorithm2e}

\usepackage[pagebackref=true,breaklinks=true,letterpaper=true,colorlinks,bookmarks=false]{hyperref}

\cvprfinalcopy 

\def\cvprPaperID{2540} 

\ifcvprfinal\fi
\begin{document}

\title{3D-MAN: 3D Multi-frame Attention Network for Object Detection}

\author{Zetong Yang$^{1\ast}$~~~~~~~~
Yin Zhou$^{2}$~~~~~~~~
Zhifeng Chen$^{3}$~~
Jiquan Ngiam$^{3}$~~~~~~
\\
$^{1}$The Chinese University of Hong Kong~~~~~~$^{2}$Waymo LLC~~~~~~$^{3}$Google Research, Brain Team\\
\vspace{-2mm}
{\small\tt tomztyang@gmail.com ~~~ yinzhou@waymo.com ~~~ \{zhifengc, jngiam\}@google.com} 
}

\maketitle

\begin{abstract}
3D object detection is an important module in autonomous driving and robotics. However, many existing methods focus on using single frames to perform 3D detection, and do not fully utilize information from multiple frames. In this paper, we present 3D-MAN: a 3D multi-frame attention network that effectively aggregates features from multiple perspectives and achieves state-of-the-art performance on Waymo Open Dataset. 3D-MAN first uses a novel fast single-frame detector to produce box proposals. The box proposals and their corresponding feature maps are then stored in a memory bank. We design a multi-view alignment and aggregation module, using attention networks, to extract and aggregate the temporal features stored in the memory bank. This effectively combines the features coming from different perspectives of the scene. 
We demonstrate the effectiveness of our approach on the large-scale complex Waymo Open Dataset, achieving state-of-the-art results compared to published single-frame and multi-frame methods.
\end{abstract}

{\let\thefootnote\relax\footnotetext{$^\ast$Work done during an internship at Google Brain.}}

\section{Introduction}

3D object detection is an important problem in computer vision as it is widely used in applications, such as autonomous driving and robotics. Autonomous driving platforms require precise 3D detection to build an accurate representation of the world, which is in turn used in downstream models that make critical driving decisions. 

\begin{figure}[bpt]
  \centering
  \includegraphics[width=0.9\linewidth]{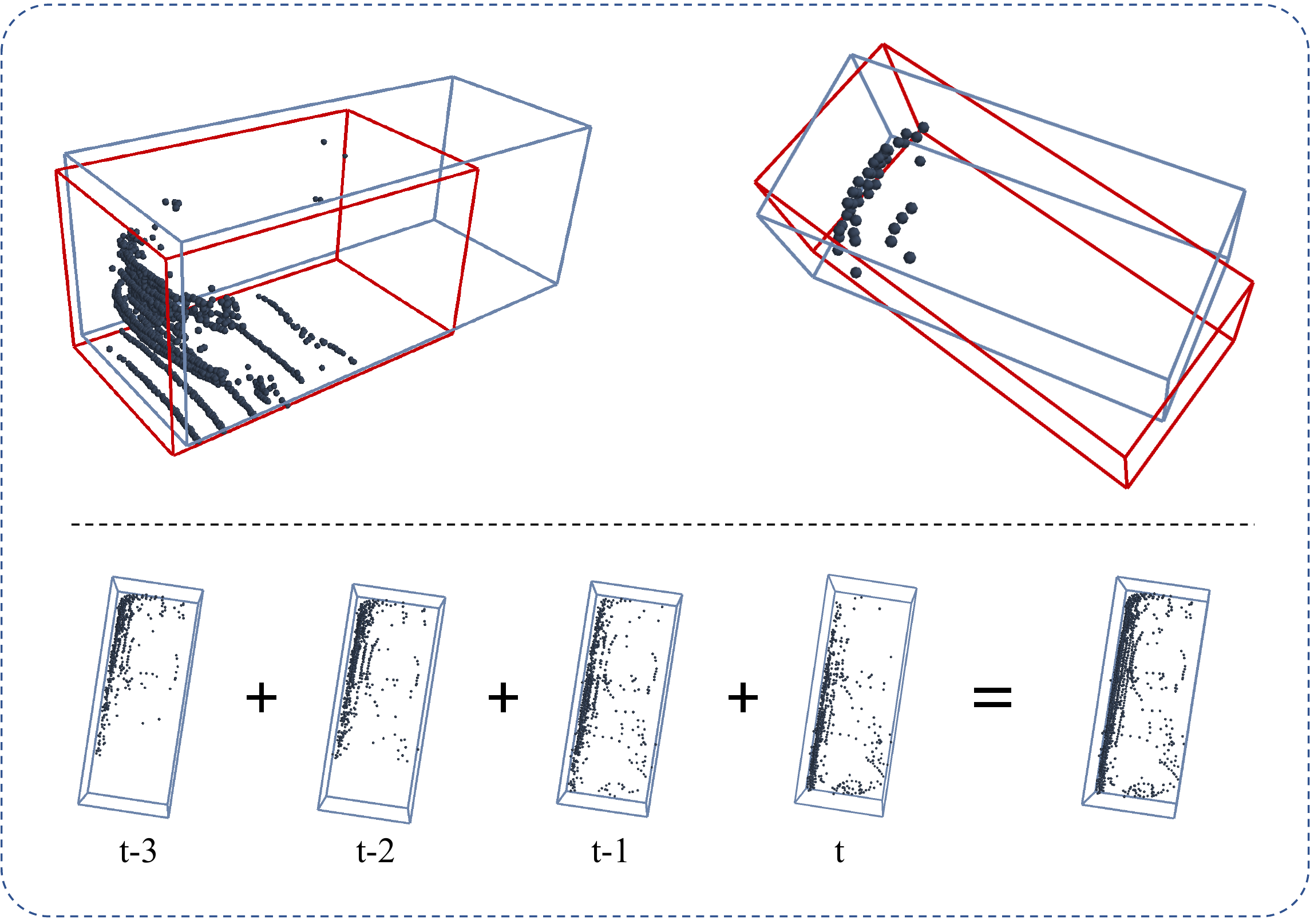}\\
  \caption{Upper row: Potential detections given LiDAR from a single frame demonstrating ambiguity between many reasonable predictions.
  Lower row: After merging the points aligned across 4 frames, there is more certainty for the correct box prediction.}
  \label{fig:ambiguity}
\end{figure}

LiDAR provides a high-resolution accurate 3D view of the world. However, at any point of time, the LiDAR sensor collects only a single perspective of the scene. It is often the case that the LiDAR points detected on an observed object correspond to only a partial view of it. Detecting these partially visible instances is an ill-posed problem because there exist multiple reasonable predictions (shown as red and blue boxes in the upper row of Figure \ref{fig:ambiguity}). These potential ambiguous scenarios can be a bottleneck for single-frame 3D detectors (Table \ref{tab:iou_compare}). 

In the autonomous driving scenario, as the vehicle progresses, the sensors pick up multiple views of the world, making it possible to resolve the aforementioned localization ambiguity. Multiple frames across time can provide different perspectives of an observed object instance. An effective multi-frame detection method should be able to extract relevant features from each frame and aggregate them, so as to obtain a representation that combines multiple perspectives (Figure \ref{fig:ambiguity}).
Research in 3D multi-frame detection has been limited due to a lack of available datasets with well-calibrated multi-frame data. Fortunately, recently released large-scale 3D sequence datasets (NuScenes \cite{nuscenes2019}, Waymo Open Dataset \cite{Waymo}) have made such data available.

\begin{table}[t]
   \centering \addtolength{\tabcolsep}{1.3mm}
   \footnotesize
   \begin{tabular}{c|c|c|c}
       \hline
       IoU threshold & 0.3 & 0.5 & 0.7 \\
       \hline
       AP (\%) & 94.72 & 88.97  & 63.27 \\
      \hline
   \end{tabular}\vspace{0.2cm}
   \caption{We vary the intersection-over-union (IoU) threshold for considering a predicted box correctly matched to a ground-truth box, and measure the performance of the PointPillars model on the Waymo Open Dataset's validation set. A lower IoU threshold corresponds to allowing less accurate boxes to match. This shows that improving the box localization could significantly improve model performance.}
   \label{tab:iou_compare}
\end{table}

A straight-forward approach to fusing multi-frame point clouds is to use point concatenation, which simply combines points across different frames together \cite{nuscenes2019}. The combined point cloud is then used as input to a single-frame detector. This approach works well for static and slow-moving objects since the limited movement implies that the LiDAR points will be mostly aligned across the frames. However, when objects are fast-moving or when longer time horizons are considered, this approach may not be as effective since the LiDAR points are no longer aligned (Table \ref{tab:Concat_comp}).

\begin{table}[t]
   \centering \addtolength{\tabcolsep}{0mm}
   \footnotesize
   \begin{tabular}{c|c|c|c|c}
       \hline
       Model & Stationary (\%) & Slow (\%) & Medium (\%) & Fast (\%) \\
       \hline
       1-frame & 60.01 & 66.64  & 65.02 & 71.90  \\
       \hline
       4-frames & 62.4 & 67.39 & \bf 66.68 & \bf 77.99 \\
       \hline
       8-frames & \bf 63.7 & \bf 67.98 & 66.29 & 72.30 \\ 
      \hline
   \end{tabular}\vspace{0.1cm}
   \caption{Velocity breakdowns of vehicle AP metrics for PointPillars models using point concatenation. For the 8-frame model, we find that its benefits come from slow-moving vehicles. Fast-moving objects no longer benefit from a large number of frames since the LiDAR points are no longer aligned across the frames. }
   \label{tab:Concat_comp}
\end{table}


As an alternative to point concatenation, Fast-and-furious \cite{FaF} attempts to fuse information across frames by concatenating at a feature map level. However, this still runs into the same challenge with misaligned feature maps for fast-moving objects and longer time horizons. Recent approaches \cite{AnLSTMApproach, 3DVID} propose using recurrent layers such as Conv-LSTM or Conv-GRU to aggregate the information across frames. It turns out that these recurrent approaches are often computationally expensive.


\vspace{10pt}\noindent\textbf{Our Approach.} 
We propose 3D-MAN: a 3D multi-frame attention network that is able to extract relevant features from past frames and aggregate them effectively. 3D-MAN has three components: (i) a fast single-frame detector, (ii) a memory bank, and (iii) a multi-view alignment and aggregation module.


The fast single-frame detector (FSD) is an anchor-free one-stage detector with a novel learning strategy. We show that a max-pooling based non-maximum suppression (NMS) algorithm together with a novel Hungarian-matching based loss is an effective method to generate high-quality proposals at real-time speeds. These proposals and the last feature map from FSD are then fed into a memory bank. The memory bank stores both predicted proposals and feature maps in previous frames so as to maintain different perspectives for each instance across frames. 

The stored proposals and features in the memory bank are finally fused together through the multi-view alignment and aggregation module (MVAA), which produces fused multi-view features for target proposals that are used to regress bounding boxes for final predictions. MVAA has two stages: a multi-view \textit{alignment} stage followed by a multi-view \textit{aggregation} stage. The alignment stage works on each stored frame independently; it uses target proposals as queries into a stored frame to extract relevant features. The aggregation stage then merges across frames for each target proposal independently. This can be viewed as a form of factorization over the attention across proposals and frames.

We evaluate our model on large-scale Waymo Open Dataset \cite{Waymo}. Experimental results demonstrate that our method outperforms published state-of-the-art single-frame methods and multi-frame methods. Our primary contributions are listed below.

\vspace{10pt}\noindent\textbf{Key Contributions.} 

\begin{itemize}\vspace{-0.05in}

\item We propose 3D-MAN: a 3D multi-frame attention network for object detection. We demonstrate that our method achieves state-of-the-art performance on the Waymo Open Dataset \cite{Waymo} and provide thorough ablation studies.

\vspace{-0.05in}
\item We introduce a novel training strategy for a fast single-frame detector method that uses max-pooling to perform non-maximum suppression and a variant of Hungarian matching to compute a detection loss.

\vspace{-0.05in}
\item We design an efficient multi-view alignment and aggregation module to extract and aggregate relevant features from multiple frames in a memory bank. This module produces features containing information from multiple perspectives that perform well for classification and bounding box regression.

\end{itemize}

\section{Related Work}

\vspace{10pt}\noindent\textbf{3D Single-frame Object Detection.}
Current 3D object detectors can be categorized into three approaches: voxel-based methods, point-based methods, and their combination. First, voxel-based methods transform via voxelization a set of unordered points into a fixed-size 2D feature map, on which convolutional neural networks (CNN) can be applied to generate detection results. Traditional approaches for voxel feature extraction rely on hand-crafted statistical quantities or binary encoding \cite{VotetoVote,YangLU18}, while recent works show that machine-learned features demonstrate favorable performance \cite{VOXELNET,lang2018pointpillars,yan2018second,MVF,shi2019part,REF:PillarNet_ECCV2020}.
Second, point-based methods \cite{yang2018ipod,shi2018pointrcnn,yang3DSSD20,StarNet,yang2019std} address detection problems by directly extracting features based on the point cloud, without an explicit discretization step. Finally, recent works have combined methods from both voxel-based and point-based feature representations \cite{PVRCNN} by using the voxel-based methods to generate proposals and the point-based methods to refine them.

\begin{figure*}[bpt]
  \centering
  \includegraphics[width=1.0\linewidth]{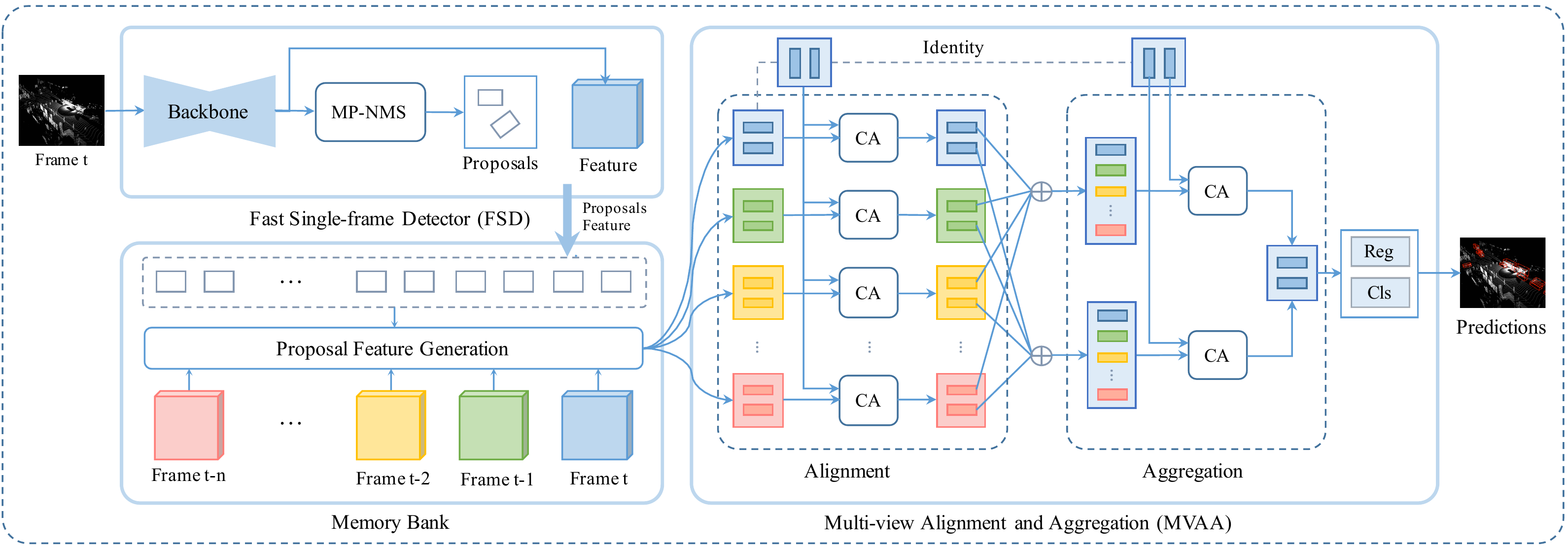}\\
  \caption{Framework for 3D-MAN: 3D multi-frame attention network. Given the point cloud for a target frame $t$, a fast single-frame detector first generates box proposals. These proposals (box parameters) with the feature map (last layer of the backbone network) are inserted into a memory bank that stores proposals and features for the last $n$ frames. We use a proposal feature generation module to extract proposal features for each stored frame. Each small rectangle box denotes a proposal and its associated features extracted in different frames. The multi-view alignment and aggregation module performs attention across proposal features from the memory bank, using the target frame as queries to extract features for classification and regression. ``MP-NMS'' and ``CA'' represent MaxPoolNMS and cross-attention respectively. During training, we use classification and regression losses applied to the FSD proposals ($L_{fsd}$), the final outputs of the MVAA network ($L_{mvaa}$), and the outputs of the alignment stage ($L_{cv}$, an auxiliary cross-view loss).}
  \label{fig:framework}
\end{figure*}

\vspace{10pt}\noindent\textbf{2D Multi-frame Object Detection.}
2D multi-frame object detection has been widely explored compared to 3D counterparts. 2D detection methods primarily focus on aligning objects in a target frame using motion and appearance features from previous frames. Relational modules with self-attention layers \cite{relation} are prevalent among these methods \cite{VOD1,VOD2,VOD3,VOD4,VOD5}. They usually take as input a target frame and multiple reference frames, from which proposals are generated per frame. Relation modules are applied to aggregate temporal features for more robust object detection. Most approaches use self-attention across all proposals in all previous frames.
In contrast, our method factorizes the attention layer to first operate independently across frames (alignment stage), and then independently across proposals (aggregation stage).

\vspace{10pt}\noindent\textbf{3D Multi-frame Object Detection.}
A straight-forward approach to multi-frame detection is to concatenate the points from different frames together \cite{nuscenes2019}. This has been demonstrated on the NuScenes dataset (improvement of $21.9 \%$ to $28.8 \%$ mAP \cite{nuscenes2019}), and we also observe improvements in our experiments (Table \ref{tab:Concat_comp}). However, as we increase the number of frames concatenated, the improvement diminishes since the LiDAR points are less likely to be aligned across longer time horizons (Table \ref{tab:Concat_comp}).
Fast-and-furious \cite{FaF} side steps aligning the points by instead concatenating the intermediate features maps. However, this approach may still result in misalignment across the feature maps for fast-moving objects and longer time horizons. Recent approaches \cite{AnLSTMApproach,3DVID} show further performance improvement by applying Conv-LSTM or Conv-GRU to fuse multi-frame information. However, the use of a single memory state that gets updated creates a potential bottleneck, and the high resolution of the feature maps make these methods computationally expensive.

\section{3D-MAN Framework}

The 3D-MAN framework (Figure \ref{fig:framework}) consists of 3 components: (i) a fast single-frame detector (FSD) for producing proposals given input point clouds, (ii) a memory bank to store features from different frames and (iii) a multi-view alignment and aggregation module (MVAA) for combining information across frames to generate final predictions.

\subsection{Fast Single-frame Detector}

\vspace{10pt}\noindent\textbf{Anchor-free Point Pillars.} We base our single-frame detector on the PointPillars architecture \cite{lang2018pointpillars} with dynamic voxelization \cite{MVF}. We start by dividing the 3D space into equally distributed pillars which are voxels of infinite height. Each point in the point cloud is assigned to a single pillar. Each pillar is then featurized using a PointNet \cite{POINTNET} producing a 2D feature representation for the entire scene, which is subsequently processed through a CNN backbone. Each location of the final layer of the network produces a prediction for a bounding box relative to the corresponding pillar center. We regress the location residuals, bounding box sizes, and orientation. A binning approach is used for predicting orientation which first classifies the orientation into one bin followed by regression of the residual from the corresponding bin center \cite{FPOINTNET,shi2018pointrcnn}.

Non-maximum suppression (NMS) is often used to post-process the detections produced by the last layer of the network for redundancy removal. It first outputs the highest scoring box and then suppresses all overlapping boxes with that box, repeating this process until all boxes are processed. However, the sequential nature of this algorithm makes it slow to run in practice when there is a large number of predictions. We use a variant of NMS that leverages max-pooling to speed up this process. MaxPoolNMS~\cite{CenterNet} uses the max pooling operation to find local peaks on the objectness score map. The local peaks are kept as predictions, while all other locations are suppressed. This process is fast and highly parallelizable. We find that this approach can be up to $6\times$ faster\footnote{Execution time for regular NMS depends on the number of output boxes desired, while MaxPoolNMS's speed is invariant the number of output boxes.} than regular NMS when dealing with about 200k predictions.


\vspace{10pt}\noindent\textbf{Hungarian Matching.} 
MaxPoolNMS is usually performed using the classification score as the ranking signal to indicate that one box is better than another. However, the classification score is a proxy metric: ideally, we want to have the highest scoring box to be the best localized box.
The ideal score map should have a single peak which corresponds to the best localized box. We propose using the Hungarian matching algorithm \cite{detr,HungMatch1} to produce such a score map.

Given a set of bounding box predictions and a set of ground-truth boxes, we compute the IoU score for each pair of them. By applying the Hungarian matching algorithm to this matrix\footnote{In practice, we add dummy boxes to the ground-truth boxes so that a one-to-one match is always produced.} of pair-wise scores, we can obtain a single match for each ground-truth box to a predicted box that maximizes the overall matching score. 
For each ground-truth box, we treat the matched predicted box as positive, and all unmatched boxes as negative. In this way, the model is encouraged to predict only one positive box per ground-truth box such that the box predicted corresponds to the highest IoU-scoring box.

It turns out that there are two challenges when using the Hungarian matching algorithm. First, the Hungarian matching algorithm is of order $O(n^3)$ and can be slow if there are a large number of predictions. Therefore, we choose to perform the Hungarian matching based-loss only after the MaxPoolNMS step. This ensures that only a few predictions remain, and enables the matching algorithm to complete quickly.

Second, the model can end up in a bad local minima by only predicting boxes which are far away from any ground-truth box (e.g., predicting boxes in locations where there are no points in the input point clouds). Consequently, these ground-truth boxes do not overlap at all with their matched prediction boxes. As a result, the model does not get any meaningful learning signals from these matches and is not able to converge to a good solution. To address this issue, we post-process the matches to reassign ground-truth boxes that have no overlap with their matched prediction box. We assign them instead to their closest pillar in the feature map, which may not be one retained by MaxPoolNMS. This encourages the model to avoid invalid assignments and converge well. 

\begin{figure}[t]
  \centering
  \includegraphics[width=0.9\linewidth]{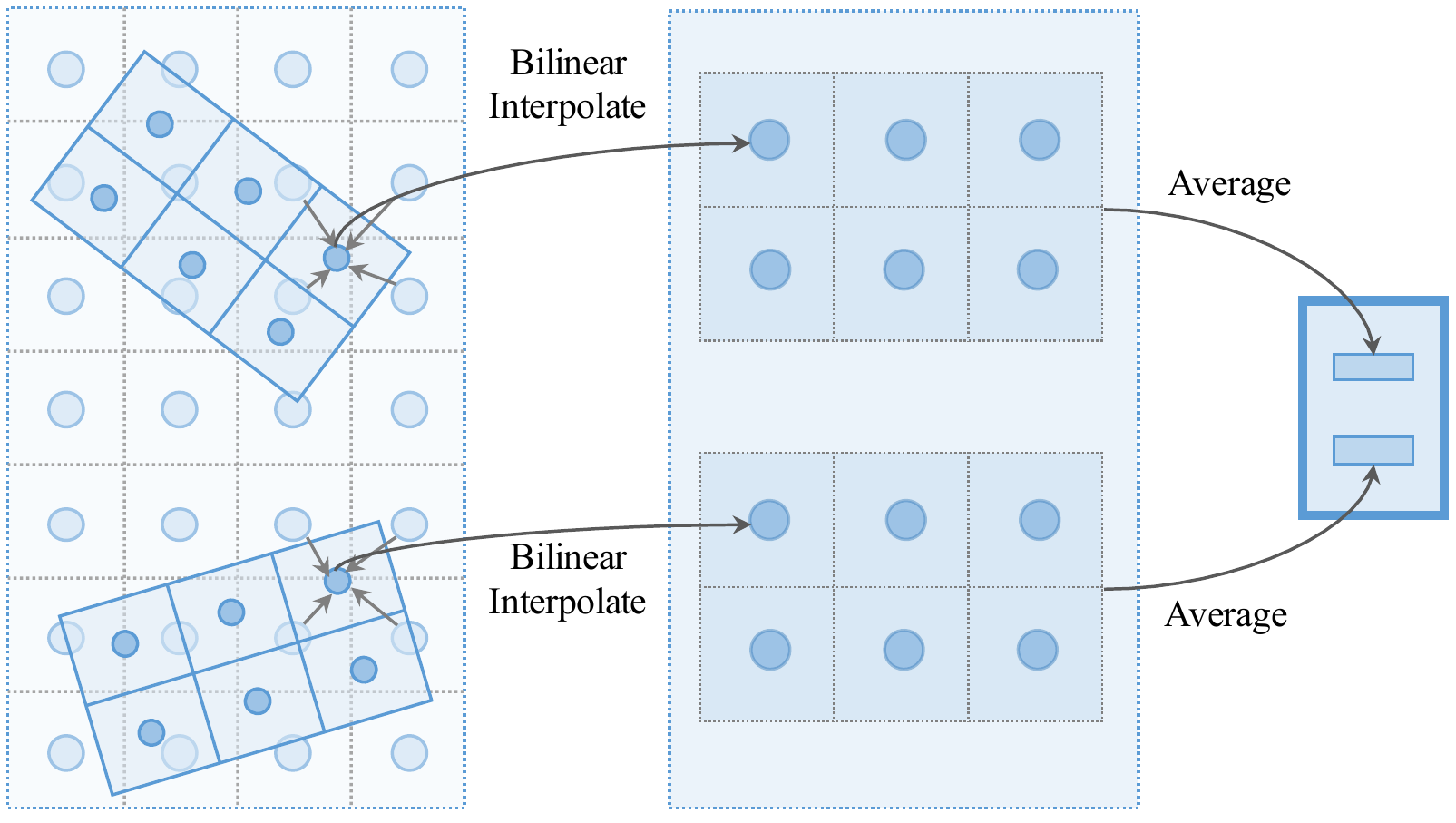}\\
  \caption{Illustration of rotated ROI feature extraction \cite{Liang2019CVPR}. We first identify key points in each proposal box and then extract features using bilinear interpolation. Averaging pooling is further used to summarize each box into a single feature vector. Note that while the figure denotes key points over $3\times2$ locations, we use $7\times7$ for vehicles and $3\times3$ for pedestrians.}
  \label{fig:pfg}
\end{figure}

\subsection{Memory Bank}

\vspace{5pt}\noindent\textbf{Memory Bank.}
We use a memory bank to store the proposals and feature maps extracted by the FSD for the last $n$ frames. When proposals and features from a new frame are added to the bank, those from the oldest frame are discarded. 



\vspace{5pt}\noindent\textbf{Proposal Feature Generation.}
To obtain features from multiple perspectives, we propose to generate proposal features for each stored frame in the memory bank as well as the target frame. We find that it is useful to use all stored proposals regardless of which frame the proposal comes from to extract features from every stored frame. This allows the model to increase its recall since an object may be missed by FSD in a single frame because of occlusion or partial observation.

For each proposal, we extract its features using a rotated ROI feature extraction approach (Figure \ref{fig:pfg}) \cite{Liang2019CVPR}. Given a proposal, we identify $K\times K\times1$ equally distributed key points with respect to the proposal box\footnote{We use $7\times7\times1$ for vehicles and $3\times3\times1$ for pedestrians.}. For each key point, we compute a feature by bilinear interpolation of its value in the feature map. Finally, we use average pooling across all the $K\times K\times1$ key points to obtain a single feature vector for the proposal. It is worth noting that this feature extraction method can be performed without correcting the entire LiDAR point cloud for ego-motion of the autonomous vehicle. This facilitates deployment in a production autonomous driving system.

The proposal features generated for the target frame will be used next in the MVAA module as the query features for the cross-attention networks, while proposal features for stored frames will be treated as keys and values. 

\subsection{Multi-view Alignment and Aggregation}

These proposal features are then sent to the multi-view alignment and aggregation module (MVAA) to be extracted and aggregated. The alignment module is applied independently for each stored frame (attention is across boxes, performed separately for each frame), while the aggregation module is applied independently for each box (attention is across time). One can view this as a factorized form of attention.

\begin{figure}[bpt]
  \centering
  \includegraphics[width=1.0\linewidth]{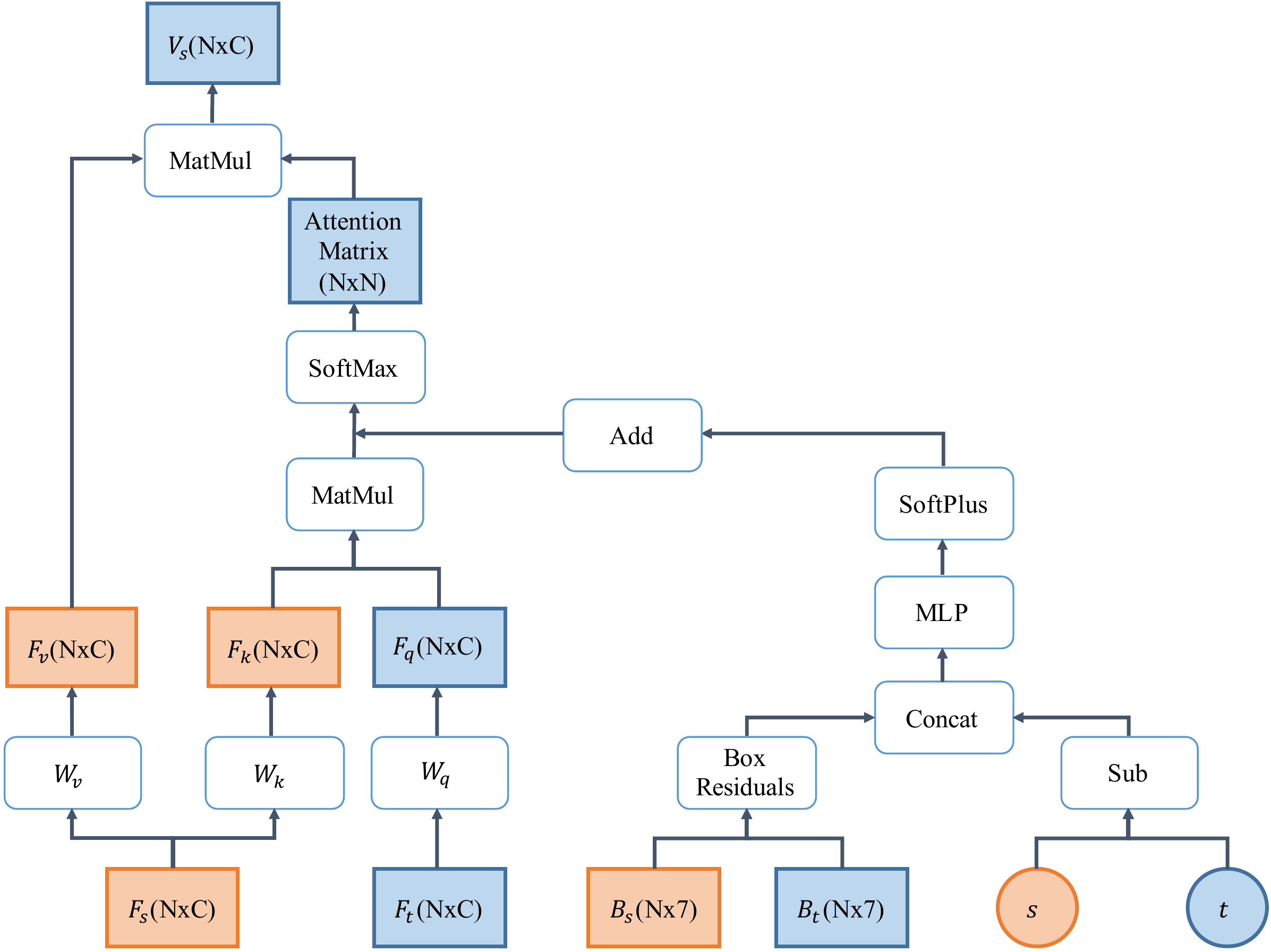}\\
  \vspace{3pt}
  \caption{Cross-attention network in the multi-view alignment module. $F_s$ and $B_s$ represent features and box parameters of proposals in a stored frame while $F_t$ and $B_t$ are those for the target frame. We use $s$ and $t$ to denote the indices of the stored frame and target frame respectively. $N$ and $C$ stand for the number of proposals and channels. ``Box residuals'' produces a pair-wise $N\times N \times 7$ tensor that
  encodes the differences in all pairs of boxes, using the same approach that is used to compute residuals for ground-truth boxes from anchor boxes  \cite{lang2018pointpillars}. $V_s$ is the output of the cross-attention network, such that each input target box has one associated output feature vector with the corresponding stored frame.}
  \label{fig:cross_attention}
\end{figure}

\vspace{10pt}\noindent\textbf{Multi-view Alignment.}
Given a new frame's proposal, the multi-view alignment module is responsible for extracting its relevant information in each previous frame separately (Figure \ref{fig:framework}, MVAA-Alignment). To achieve this goal, the alignment stage has to figure out how to relate the identities of the proposals in the new frame to those in the stored frames.
A naive approach could use nearest neighbor matching or maximum IoU overlap. However, when an instance is fast-moving or close to any other instance, there will often be ambiguity in the appropriate assignment. Furthermore, the naive approach does not learn interactions between the new proposal and other objects in previous frames that could provide contextual information.

We propose using a cross-attention network (Figure \ref{fig:cross_attention}) to learn how to relate the new frame proposals to those of stored frames. This network could potentially learn to align the proposal identities and also model interactions across objects. Specifically, we apply projection layers to encode the new frame proposal features $F_t$ as well as stored proposal features $F_s$ so as to compute projected queries $F_q$, keys $F_k$ and values $F_v$. These are used to compute an attention matrix. We further provide temporal and spatial information to the attention matrix through encoding 
the relative frame index and box residuals between all pairs of the query and stored boxes. The cross-attention network is applied between the target frame and each stored frame independently with shared parameters, generating a feature vector for each target proposal ($V_s$) from each stored frame.

\vspace{10pt}\noindent\textbf{Cross-view Loss.}
The alignment stage of MVAA is designed to extract features from each stored frame that are most relevant to each target proposal. To encourage the extracted features to be a relevant representation, we employ an auxiliary loss that encourages the extracted features to contain sufficient information to predict the corresponding ground-truth bounding box associated with the target proposal. Concretely, we add separate classification and regression heads that use each extracted feature vector of the alignment stage to predict the box residuals between target proposal and its corresponding ground-truth box.






\vspace{10pt}\noindent\textbf{Multi-view Aggregation.}
After the alignment module, each proposal in the target frame will have an associated feature for each stored frame. The multi-view aggregation layer (Figure \ref{fig:framework}, MVAA-Aggregation) is responsible for combining these features from different perspectives together to form a single feature for each proposal.
Concretely, we use the new frame's proposal features as the attention query inputs, and its corresponding extracted features in previous frames as the keys and values.

We note that the aggregation module can enable the network to be robust to newly appearing objects. If an object appears for the first time in a new frame, the model can compute an attention matrix that will only focus on the new frame and ignore the past frames since they are not relevant.

\vspace{10pt}\noindent\textbf{Box Prediction Head.}
After MVAA, we have an updated feature for each proposal in the new frame. We regress objectness scores and box parameters from this feature representation. For the objectness score, we follow \cite{PVRCNN} and treat the IoU between proposals and their corresponding ground-truth bounding boxes as the classification target, with the sigmoid cross-entropy loss. The box parameter targets are encoded as residuals \cite{lang2018pointpillars,VOXELNET} and trained with a smooth-$L1$ loss. The same formulations are used for the cross-view loss.

\subsection{Losses}
We minimize the total loss consisting of a fast single-frame detector (FSD) loss $L_{fsd}$, a multi-view prediction loss $L_{mvaa}$, and a cross-view loss $L_{cv}$ with equal loss weights.

\begin{equation} \label{eq:total_loss}
\begin{aligned}
L_{total} = L_{fsd} + L_{mvaa} + L_{cv}
\end{aligned}
\end{equation}

The same formulation for detection loss $L_{det}$ is used in these three losses. This includes a objectness loss $L_{obj}$ and a regression loss $L_{reg}$. 

\begin{equation} \label{eq:pred_loss}
\begin{aligned}
L_{det} = \frac{1}{|C|} \sum_{i\in C} L_{obj} +\frac{1}{|R|} \sum_{i\in R}L_{reg}
\end{aligned}
\end{equation}

$C$ represents the set of locations where we predict an objectness score. For $L_{fsd}$, this corresponds to the remaining pillars after MaxPoolNMS, while for $L_{mvaa}$ and $L_{cv}$, this corresponds to the proposals after FSD (specifically, those that remain after MaxPoolNMS). For the objectness loss in $L_{mvaa}$ and $L_{cv}$, we use the IoU overlap between the proposal and its assigned ground-truth as the target. For $L_{fsd}$, the output of the Hungarian matching is used to determine positive and negative assignments for the objectness loss. 

$R$ represents the set of locations which are associated with a ground-truth box. For all losses ($L_{fsd}$, $L_{mvaa}$, and $L_{cv}$), these are the matched boxes from Hungarian matching. 
For the regression losses, we use a smooth-$L1$ loss as the supervision of regressing the x, y, z center location residuals, and their corresponding dimensions. For orientation, we use a binning orientation loss \cite{FPOINTNET,shi2018pointrcnn}. The model is expected to predict an angle bin first, followed by a residual from the bin center. We use 12 bins for $L_{fsd}$ and 1 bin for $L_{mvaa}$ and $L_{cv}$.

The cross-view identity loss $L_{cv}$ is computed across all the outputs of the multi-view alignment stage, and averaged across all instances.

\section{Experiments}

We evaluate our method on Waymo Open Dataset \cite{Waymo}, a large scale 3D object detection dataset. There are a total of 1150 sequences divided into 798 training, 202 validation, and 150 testing examples. Each sequence consists of about 200 frames at a frame rate of 10 Hz, where each frame includes a LiDAR point cloud and labeled 3D bounding boxes for vehicles, pedestrians, cyclists and signs. We evaluate our model and compare it with other methods using average precision (AP) and Average Precision Weighted by Heading (APH).

\subsection{Implementation Details}
\vspace{4pt}\noindent\textbf{Hyperparameters.}
Given the input point cloud in a target frame, we first set the detection range as $[-76.8m, 76.8m]$ for x and y axes and $[-2m,4m]$ for the z-axis. We equally split this 3D range into $[512, 512]$ pillars among x and y axes respectively, following PointPillars \cite{lang2018pointpillars}.
For the MaxPoolNMS applied in FSD, we use a max-pooling kernel size of $[7, 7]$ for vehicles and $[3, 3]$ for pedestrians, with a stride of $[1,1]$. After MaxPoolNMS, a set of $128$ proposals per frame are passed to the memory bank.

\vspace{4pt}\noindent\textbf{Network Architectures.}
In our proposed FSD, we use the same backbone network illustrated in PointPillars \cite{lang2018pointpillars}. The channel dimension $C$ of the last feature map and proposal features is 384. For encoding the frame index and relative box residuals, we apply a 2-layer perceptron (MLP) networks with $C$ output channels for the first layer, and $1$ output for the second layer. These are used in the cross-attention layers of the MVAA module. In the prediction head, we first apply a 2-layer MLP network with $C$ output channels to embed the aggregated multi-view features. These embeddings are transformed with two prediction branches for classification and regression.

\vspace{4pt}\noindent\textbf{Training Parameters.}
Our network is trained end-to-end using the ADAM \cite{AdamOptimizer} optimizer for a total number of 50 epochs with an initial learning rate of $0.0016$ and a batch size of 32.
We apply exponential decay to anneal the learning rate, starting at 5 epochs until 45 epochs. During training, we apply random flip and random rotation as our only data augmentation methods.

\begin{table*}[t]
   \centering 
   \setlength{\tabcolsep}{3mm}
   \footnotesize
   \begin{tabular}{c|c|cccc|cccc}
       \hline
       \multicolumn{1}{c|}{ \multirow{2}{*}{Difficulty}} &
       \multicolumn{1}{c|}{ \multirow{2}{*}{Method}} & \multicolumn{4}{c|}{ \multirow{1}{*}{3D AP (IoU=0.7)}} & \multicolumn{4}{c}{ \multirow{1}{*}{3D APH (IoU=0.7)}} \\ \cline{3-10}
       {} & {} &
       \multicolumn{1}{c}{ \multirow{1}{*}{Overall}} & 
       \multicolumn{1}{c}{ \multirow{1}{*}{0-30m}} &
       \multicolumn{1}{c}{ \multirow{1}{*}{30-50m}} & 
       \multicolumn{1}{c|}{ \multirow{1}{*}{50m-Inf}} &
       \multicolumn{1}{c}{ \multirow{1}{*}{Overall}} &
       \multicolumn{1}{c}{ \multirow{1}{*}{0-30m}} &
       \multicolumn{1}{c}{ \multirow{1}{*}{30-50m}} & 
       \multicolumn{1}{c}{ \multirow{1}{*}{50m-Inf}} \\
       \hline
       \hline
       \multirow {10}{*}{LEVEL\_1} & StarNet \cite{StarNet}& 55.11 & 80.48 & 48.61 & 27.74 & 54.64 & 79.92 & 48.10 & 27.29 \\
       {} & PointPillars \cite{lang2018pointpillars} & 63.27 & 84.90 & 59.18 & 35.79 & 62.72 & 84.35 & 58.57 & 35.16 \\
       {} & MVF \cite{MVF} & 62.93 & 86.30 & 60.02 & 36.02 & - & - & - & - \\
       {} & AFDet \cite{AFDet} & 63.69 & 87.38 & 62.19 & 29.27 & - & - & - & -\\
       {} & RCD \cite{bewley2020range} & 68.95 & 87.22 & 66.53 & 44.53 & 68.52 & 86.82 & 66.07 & 43.97 \\
       {} & PV-RCNN \cite{PVRCNN} & 70.30 & 91.92 & 69.21 & 42.17 & 69.49 & 91.34 & 68.53 & 41.31 \\
       {} & 3D-MAN (Ours) & 69.03 & 87.99 & 66.55 & 43.15 & 68.52 & 87.57 & 65.92 & 42.37 \\
       \cline{2-10}
       {} & PointPillars$^\ast$ \cite{lang2018pointpillars} & 65.41 & 85.58 & 61.51 & 39.51 & 64.88 & 85.02 & 60.95 & 38.91 \\
       {} & ConvLSTM$^\ast$ \cite{AnLSTMApproach} & 63.6 & - & - & - & - & - & - & - \\
       {} & 3D-MAN$^\ast$ (Ours) & \bf74.53 & \bf92.19 & \bf72.77 & \bf 51.66 & \bf 74.03 & \bf 91.76 & \bf 72.15 & \bf 51.02 \\
       \hline
       \hline
       \multirow {6}{*}{LEVEL\_2} & StarNet \cite{StarNet} & 48.69 & 79.67 & 43.57 & 20.53 & 48.26 & 79.11 & 43.11 & 20.19 \\
       {} & PointPillars \cite{lang2018pointpillars} &  55.18 & 83.61 & 53.01 & 26.73 & 54.69 & 83.08 & 52.46 & 26.24 \\
       {} & PV-RCNN \cite{PVRCNN} & 65.36 & 91.58 & 65.13 & 36.46 & 64.79 & 91.00 & 64.49 & 35.70 \\
       {} & 3D-MAN (Ours) & 60.16 & 87.10 & 59.27 & 32.69 & 59.71 & 86.68 & 58.71 & 32.08 \\
       \cline{2-10}
       {} & PointPillars$^\ast$ \cite{lang2018pointpillars} & 57.28 & 84.31 & 55.41 & 29.71 & 56.81 & 83.79 & 54.90 & 29.24 \\  
       {} & 3D-MAN$^\ast$ (Ours) & \bf67.61 & \bf92.00 & \bf67.20 & \bf41.38 & \bf67.14 & \bf91.57 & \bf66.62 & \bf40.84 \\
      \hline
   \end{tabular}\vspace{0.2cm}
   \caption{3D AP and APH Results on Waymo Open Dataset validation set for class Vehicle. $^\ast$Methods utilize multi-frame point clouds for detection. We report PointPillars \cite{lang2018pointpillars} based on our own implementation, with and without point concatenation. Difficulty levels are defined in the original dataset\cite{Waymo}. \label{tab:mainwaymo}}
\end{table*}

\begin{table}[t]
   \centering \addtolength{\tabcolsep}{0mm}
   \footnotesize
   \begin{tabular}{c|c|c|c|c}
       \hline
       \multicolumn{1}{c|}{ \multirow{2}{*}{Method}} &
       \multicolumn{2}{c|}{ \multirow{1}{*}{LEVEL\_1}} & 
       \multicolumn{2}{c}{ \multirow{1}{*}{LEVEL\_2}} \\ \cline{2-5}
       {} &
       \multicolumn{1}{c|}{ \multirow{1}{*}{3D AP}} & 
       \multicolumn{1}{c|}{ \multirow{1}{*}{3D APH}} &
       \multicolumn{1}{c|}{ \multirow{1}{*}{3D AP}} & 
       \multicolumn{1}{c}{ \multirow{1}{*}{3D APH}} \\ \cline{1-5}
       \hline
       StarNet \cite{StarNet} & 68.32 & 60.89 & 59.32 & 52.76 \\
       PointPillars \cite{lang2018pointpillars} & 68.88 & 56.57 & 59.98 & 49.14 \\
       MVF \cite{MVF} & 65.33 & - & - & - \\
       3D-MAN (Ours) & \bf 71.71 & \bf 67.74 & \bf 62.58 & \bf 59.04 \\
      \hline
   \end{tabular}\vspace{0.2cm}
   \caption{3D AP and APH Results on Waymo Open Dataset \textit{validation} set for class Pedestrian.}
   \label{tab:waymo_ped}
\end{table}

\vspace{4pt}\noindent\textbf{Utilizing a large number of frames.}
We enable 3D-MAN to exploit a large number of frames by combining it with point concatenation. Our best model uses 16 frames split into 4 windows of 4 frames. The point clouds in each window are concatenated together and used as input to the FSD. Each window thus becomes an entry in the memory bank, and the model is expected to produce predictions for only the last frame. This utilizes point concatenation for when movement is small with nearby frames and MVAA for large movement across a longer time range. We provide further ablation studies with varying sizes of input frames in Section \ref{sec:frame-number-ablation}.

\subsection{Main Results}

\vspace{4pt}\noindent\textbf{Waymo Validation Set.}
We compare our method with published state-of-the-art single-frame and multi-frame methods on the Waymo validation set on class Vehicle (Table \ref{tab:mainwaymo}) and class Pedestrian (Table \ref{tab:waymo_ped}). We first compare the performance between our model with and without multi-frame inputs (Table \ref{tab:mainwaymo}). When the model has access to 16 stored frames, the overall 3D AP (LEVEL\_1) is improved by $5.50 \%$ on vehicles labeled as LEVEL\_1 difficulty, illustrating the effectiveness of our approach.

3D-MAN outperforms the current best published method (PV-RCNN \cite{PVRCNN}) by $3.56 \%$ (30-50m range) and $9.49 \%$ ($>$50m range) AP (LEVEL\_1) on vehicles. At these further ranges, objects are often partially visible, where having more information from different perspectives could help. 
These improvements show that our model is able to effectively combine the information across multiple views to generate more accurate 3D predictions. Moreover, compared to existing multi-frame models, 3D-MAN also outperforms them by a large margin. Our method achieves a better 3D AP than the recently published Conv-LSTM method~\cite{AnLSTMApproach} by $10.93 \%$ on vehicle detection.
For pedestrian detection, 3D-MAN also achieves the best performance (Table \ref{tab:waymo_ped}). 

\begin{table}[t]
   \centering \addtolength{\tabcolsep}{0mm}
   \footnotesize
   \begin{tabular}{c|c|c|c|c}
       \hline
       \multicolumn{1}{c|}{ \multirow{2}{*}{Method}} &
       \multicolumn{2}{c|}{ \multirow{1}{*}{Vehicle}} & 
       \multicolumn{2}{c}{ \multirow{1}{*}{Pedestrians}} \\ \cline{2-5}
       {} &
       \multicolumn{1}{c|}{ \multirow{1}{*}{3D AP}} & 
       \multicolumn{1}{c|}{ \multirow{1}{*}{3D APH}} &
       \multicolumn{1}{c|}{ \multirow{1}{*}{3D AP}} & 
       \multicolumn{1}{c}{ \multirow{1}{*}{3D APH}} \\ \cline{1-5}
       \hline
       SECOND \cite{yan2018second} & 50.11 & 49.63 & - & - \\
       StarNet \cite{StarNet} & 63.51 & 63.03 & 67.78 & 60.10 \\
       PointPillars \cite{lang2018pointpillars} & 68.62 & 68.08 & 67.96 & 55.53 \\
       SA-SSD \cite{He_2020_CVPR} & 70.24 & 69.54 & 57.14 & 48.82 \\
       RCD \cite{bewley2020range} & 71.97 & 71.59 & - & - \\
       3D-MAN (Ours) & \bf 78.71 & \bf 78.28 & \bf 69.97 & \bf 65.98 \\
      \hline
   \end{tabular}\vspace{0.2cm}
   \caption{3D AP and APH Results on Waymo Open Dataset \textit{testing} set for class Vehicle and Pedestrain among LEVEL\_1 difficulty objects. Metric breakdowns for our model is available on the \href{https://waymo.com/open/challenges/entry/?timestamp=1616565477179731&challenge=DETECTION_3D&email=jngiam@google.com}{Waymo challenge leaderboard}.}
   \label{tab:waymo_test}
\end{table}

\vspace{4pt}\noindent\textbf{Waymo Testing Set.}
We also evaluate our model on Waymo testing set through a test server submission. For vehicle detection (Table \ref{tab:waymo_test}), 3D-MAN achieves $78.71 \%$ AP and $78.28$ APH, outperforming RCD \cite{bewley2020range} by $6.74 \%$ and $6.69 \%$ respectively, which is currently the best published method among results generated by a single model (not using any ensemble methods). 


\subsection{Ablation Studies}
We conduct all our ablation studies only for the vehicle class, and report LEVEL\_1 difficulty results based on a subset of the full validation set. We created a mini-validation set by uniformly sampling $10\%$ of the full validation set. This results in a dataset that allows us to experiment significantly faster. We note that there is a negligible performance gap between the mini-validation and full validation set: for example, our best model obtains $74.3\%$ on the mini-validation set versus $74.5\%$ on the full validation set.

\begin{table}[t]
   \centering \addtolength{\tabcolsep}{0mm}
   \footnotesize
   \begin{tabular}{c|c|c|c}
       \hline
       & Mask & Centeredness & Hungarian Matching \\
       \hline
       Ped. (\%) & 64.7 & 67.1 & \bf 70.2 \\
       \hline
       Veh. (\%) & 44.5 & 63.7 & \bf 64.8 \\
      \hline
   \end{tabular}\vspace{0.2cm}
   \caption{Mini-validation AP comparison among different ground-truth assignment strategies using FSD for both Pedestrian and Vehicle classes.}
   \label{tab:hung_matching}
\end{table}

\vspace{10pt}\noindent\textbf{Hungarian Matching.}
We compare the performance of using different assignment strategies in FSD, including the mask strategy, centeredness strategy and Hungarian matching strategy (Table \ref{tab:hung_matching}). 
The mask strategy \cite{FCOS,Yangze2019Reppoint} assigns interior pillars of any valid object positive and all other pillars negative. However, this can lead to a discrepancy between classification score and localization accuracy. Our experiments show that this performs the least well. 
Centeredness strategy \cite{yang3DSSD20,FCOS,CenterNet} encourages pillars with closer distance to the instance center to have a higher classification score. However, the pillar in the center may not always draw the best localization prediction in point clouds: the LiDAR points often are on the surface of the vehicle and not in the interior. We find centeredness to perform better than mask, but worse than our proposed Hungarian matching approach.
FSD achieves the highest AP with the Hungarian matching strategy, which validates our approach.

\begin{table}[t]
   \centering \addtolength{\tabcolsep}{0mm}
   \footnotesize
   \begin{tabular}{c|c|c|c|c}
       \hline
       Method & Baseline & Concat & Relation & MVAA \\
       \hline
       AP ($\%$) & 68.2 & 70.5 & 70.1 & \bf 72.5 \\
      \hline
   \end{tabular}\vspace{0.2cm}
   \caption{Mini-validation AP comparison on class Vehicle among different multi-frame fusion approaches.}
   \label{tab:mvf_effect}
\end{table}

\vspace{10pt}\noindent\textbf{Multi-frame Approaches.}
We compare our method to other multi-frame approaches (Table \ref{tab:mvf_effect}), including the point concatenation approach and a self-attention approach across all previously detected boxes. 
Multi-frame models in the comparison have $4$ frames as input and are expected to predict bounding boxes for only the last frame. We also perform ego-motion pose correction to map points from the earlier frames to the pose of the last frame. 

The \textit{Baseline} model is our \textit{single-frame} two-stage model, which applies FSD to generate proposals with features and deploys box prediction head with an MLP network to refine these proposals. It achieves $68.2 \%$ AP on Vehicle and provides a baseline to compare the multi-frame models against. 

In the \textit{Concat} approach, points across all frames are combined together, and the merged point cloud is used as input to the \textit{Baseline} model. This improves upon the baseline by $2.3\%$ AP. 
The \textit{Relation} approach first extracts box proposals from multiple frames and then uses a self-attention network on \textit{all} past proposals directly to produce a prediction. This performs better than the \textit{Baseline} but worse than the \textit{Concat} model.

Our approach (MVAA) performs the best, outperforming the \textit{Concat} approach and \textit{Relation} approach by $2.0 \%$ and $2.4 \%$ respectively.


\begin{table}[t]
   \centering \addtolength{\tabcolsep}{0mm}
   \footnotesize
   \begin{tabular}{c|c|c|c}
       \hline
       Supervision Method & None & Correspondence Loss & CV Loss \\
       \hline
       AP ($\%$) & 70.7 & 71.4 & \bf 72.5 \\
      \hline
   \end{tabular}\vspace{0.2cm}
   \caption{Mini-validation AP comparison on class Vehicle using different auxiliary losses with the MVAA alignment stage.}
   \label{tab:aux_loss}
   \vspace{-0.2cm}
\end{table}

\vspace{10pt}\noindent\textbf{Cross-view loss.}
\label{sec:cross-view-results}
We find it useful to have an auxiliary cross-view loss to encourage the model to propagate relevant features in the alignment stage of MVAA. To evaluate the effectiveness of the cross-view loss, we compare it to not having an auxiliary loss and also an alternative auxiliary correspondence loss. The correspondence loss encourages elements of the attention matrix (of the alignment stage in MVAA) to be close to 1 if the query proposal matches the instance of the corresponding stored proposal, and zero otherwise.
We compare these approaches for the auxiliary loss (Table \ref{tab:aux_loss}), and find that using the cross-view loss outperforms having no auxiliary loss by $1.8 \%$ and using the correspondence loss by $1.1 \%$.

\begin{table}[t]
   \centering \addtolength{\tabcolsep}{0.5mm}
   \footnotesize
   \begin{tabular}{c|c|c|c|c|c|c}
       \hline
       Frames & 1 & 4 & 7 & 10 & 13 & 16 \\
       \hline
       AP (\%) & 68.2 & 72.5 & 73.4 & 73.5 & 73.8 & 74.3 \\
      \hline
   \end{tabular}\vspace{0.2cm}
   \caption{Mini-validation AP comparison for different number of input frames to the 3D-MAN model. All models are expected to predict only the last frame. Models with 7, 10, 13, and 16 frames use concatenated points (over windows of 4 frames) as input, with different amount of overlaps between adjacent windows.}
   \label{tab:frame_ablation}
\end{table}

\vspace{10pt}\noindent\textbf{Varying number of input frames.}
\label{sec:frame-number-ablation}
We further compare our model's performance on different number of available frames (Table \ref{tab:frame_ablation}). In order to draw a fair comparison between models with 7 through 16 frames, we fix the computation by using point concatenation over windows of 4 frames, with different degrees of overlaps between windows (similar to strides in convolution windows). We find that our model steadily improves as it has access to more input frames corresponding to longer time horizons.

\begin{table}[t]
   \centering \addtolength{\tabcolsep}{0mm}
   \footnotesize
   \begin{tabular}{c|c|c|c|c}
       \hline
       Model & Stationary (\%) & Slow (\%) & Medium (\%) & Fast (\%) \\
       \hline
       4-frames & 69.5 & 68.6 & 67.1 & 78.3 \\
       \hline
       16-frames & \bf 73.2 & \bf 70.4 & \bf 68.9 & \bf 79.2 \\ 
      \hline
   \end{tabular}\vspace{0.2cm}
   \caption{Velocity breakdowns of vehicle AP metrics for 3D-MAN with varying number of input frames.}
   \label{tab:vel_breakdown}
   \vspace{-0.2cm}
\end{table}

\vspace{10pt}\noindent\textbf{Velocity breakdowns.}
We also compare our model's performance across different velocity breakdowns. Recall that the baseline multi-frame PointPillars model performance degrades when using 8-frames versus 4-frames (Table \ref{tab:Concat_comp}). Conversely, our model demonstrates an improvement when we increase the number of frames from 4 to 16 (Table \ref{tab:vel_breakdown}). This shows that our approach is able to benefit fast-moving vehicles.

\section{Conclusion}
In this paper, we present a novel 3D object detection method, 3D-MAN, which utilizes attention networks to extract and aggregate features across multiple frames. We introduce a fast single-frame detector that utilizes a Hungarian matching strategy to align the objectness score with the best localized box. We show how the outputs of the single-frame detector can be used with a memory bank and a novel multi-view alignment and aggregation module to fuse the information from multiple frames together. Our method is effective across long time horizons and obtains state-of-the-art performance on a challenging large scale dataset.

{\small
\bibliographystyle{ieee_fullname}
\bibliography{egbib}
}

\end{document}